\documentclass{article}
\usepackage{booktabs}
\usepackage{float}

\usepackage[preprint]{corl_2026} 

\usepackage{amsmath}
\usepackage{amssymb}
\usepackage{array}
\usepackage{graphicx}
\usepackage{capt-of}
\usepackage{enumitem}
\usepackage{xspace}
\newcommand{\name}{\textsc{ReForce}\xspace}
\usepackage{hyperref}
\usepackage{wrapfig}

\title{\name: Learning Force-aware Retargeting for Dexterous Manipulation}

\author{
  Yuhang Wu,
  Lingqi Zeng,
  Changwei Jing,
  Jianglong Ye$^{\dagger}$,
  Xiaolong Wang$^{\dagger}$
  \\[3pt]
  UC San Diego
  \\[2pt]
  \url{https://wuyuhang-eai.github.io/reforce/}
  \\[2pt]
  {\small $\dagger$ Equal advising.}
}

\begin{document}
\maketitle
\vspace{-3.0em}
\begin{center}
    \includegraphics[width=\linewidth]{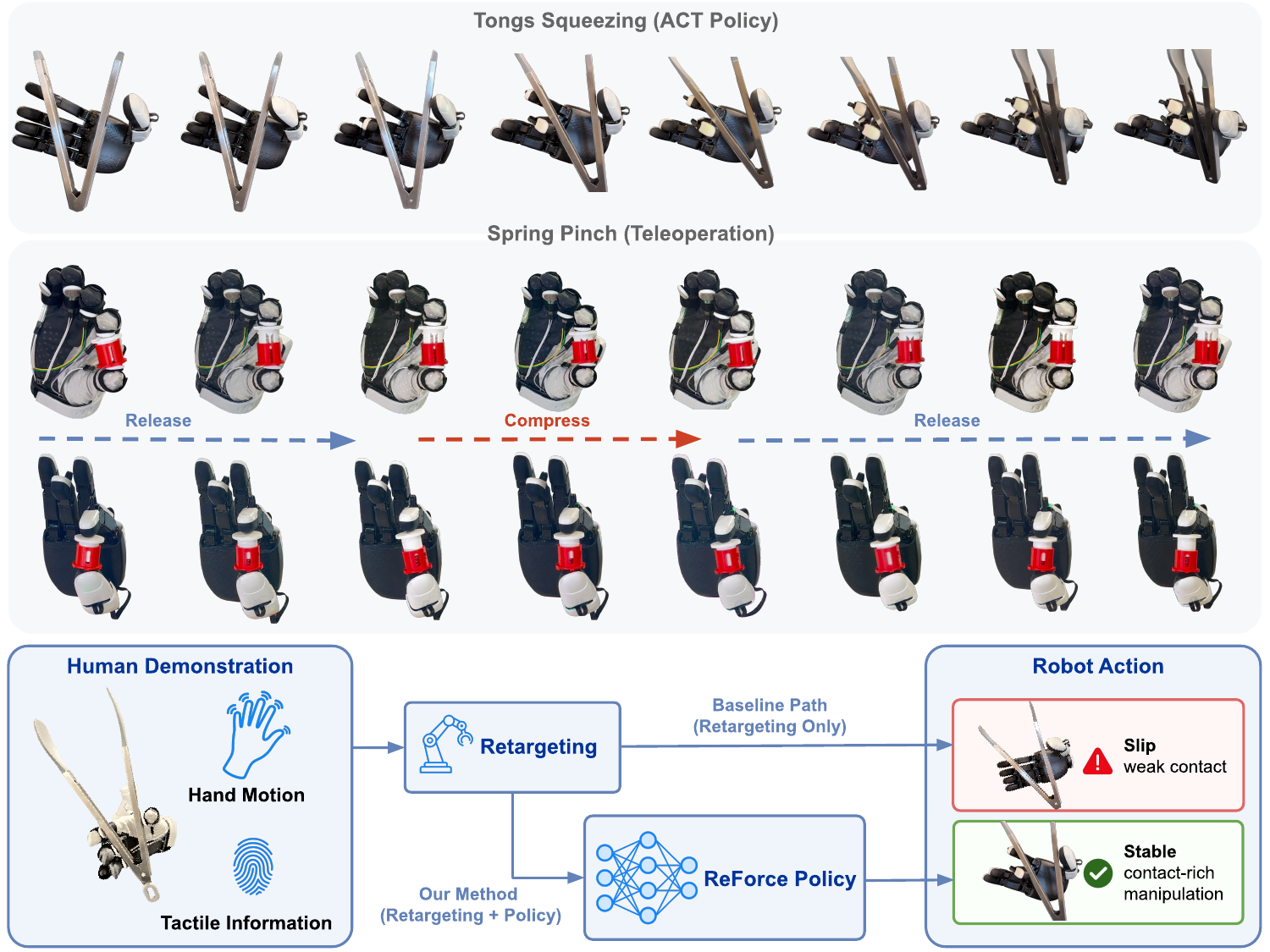}
    \vspace{-1.0em}
    \captionof{figure}{\textbf{Force-aware retargeting overview.} Human-guided motion and contact references specify the intended interaction, while \name uses robot-side fingertip force feedback to adapt the reference into dexterous-hand commands during contact-rich execution.}
    \label{fig:teaser}
\end{center}

\begin{abstract}
    Human demonstrations offer a scalable data source for dexterous manipulation, but transferring them to robot actions remains challenging due to the embodiment gap.
    Today's retargeting is mostly kinematic, yet manipulation is decided by \textit{force}, which governs how the hand interacts with the object and how the object moves.
    In this paper, we present \name, a \textbf{Force}-aware \textbf{Re}targeting method that turns human motion and forces into robot actions that reproduce the intended contact.
    \name predicts a residual on the kinematically retargeted action to reach the desired force, using a general force tracker trained on large-scale simulation interactions.
    It supports both online force-aware teleoperation and offline data translation.
    In simulation and on real hardware, \name achieves lower force-tracking error and stronger multi-finger contact engagement on contact-rich tasks such as paper-cup grasping and tongs manipulation.
\end{abstract}



\section{Introduction}

Human data is a rich source for learning dexterous manipulation~\citep{grauman2022ego4d,grauman2024egoexo4d,qiu2025humanpolicy,zheng2026egoscale,punamiya2026egoverse,li2025maniptrans,kareer2025emergence,luo2025being}, but transferring it to robot actions suffers from the embodiment gap. How to better retarget human demonstrations to deployable robot actions remains an open question. Most of today's retargeting algorithms~\citep{handa2020dexpilot,qin2023anyteleop} are kinematic only: an optimizer aligns human joint positions to their robot counterparts. Such objectives ignore geometry and contact dynamics, so their output is a coarse motion match; in practice it serves only to pre-train or co-train policies that still rely on robot data~\citep{qiu2025humanpolicy,luo2025being,zheng2026egoscale}, or runs with a human teleoperator closing the gap. Manipulation, however, is decided by \textit{force}: force determines how the hand interacts with the object, and whether a grasp holds, crushes, or slips~\citep{xie2024justaddforce,hou2025adaptive,xue2025reactivediffusion,adeniji2025feel}.

Recent methods~\citep{li2025maniptrans,mandi2026dexmachina,pan2025spider,zhu2026chord} therefore make retargeting force-aware: given a tracked hand-object demonstration, they adopt reinforcement learning or sampling in simulation to search for robot trajectories that reproduce its contact forces. These methods share two limitations: (i) each run needs a digital twin, a simulation-ready object model with accurate hand and object pose, unavailable for an arbitrary object; and (ii) each trajectory is optimized against a recorded dataset, so they cannot run during online teleoperation, which is still the main source of high-quality data.

We present \name, a \textbf{Force}-aware \textbf{Re}targeting method that turns human demonstrations into robot actions which reproduce the intended contact. Given motion and forces from a human demonstrator, \name produces robot actions that reproduce the intended contact, and transfers zero-shot to force-sensitive tasks such as grasping a paper cup or using tongs. To build it, we collect large-scale hand-object trajectories in simulation, recording both joint positions and contact forces, and train a closed-loop force tracker on them. The tracker predicts a residual on top of the kinematically retargeted action so that the robot reaches the desired force, in both online teleoperation and offline data translation.

We evaluate \name in both simulation and the real world, on force-sensitive tasks such as paper-cup grasping and tongs manipulation, under online teleoperation and offline translation settings. It achieves lower force-tracking error, while reducing severe missing-contact failures.

In summary, our contributions are threefold:
\begin{itemize}[leftmargin=*,itemsep=2pt,topsep=0pt,parsep=0pt]
    \item We propose general force tracking, a retargeting paradigm that adapts kinematic retargeting using online force feedback and requires no digital twin.
    \item We collect a large-scale simulated dataset of hand-object trajectories with paired joint positions and contact forces, and train a single force tracker on it.
    \item We evaluate in simulation and on real hardware, showing improved force tracking under both online teleoperation and offline references.
\end{itemize}

\section{Related Work}
\label{sec:related}

\paragraph{Learning from human demonstration.}
Human demonstrations provide task intent and temporal structure for robot manipulation, and recent imitation-learning methods and robot datasets have made them effective sources for trajectory generation~\citep{zhao2023aloha,shafiullah2022bet,chi2023diffusionpolicy,brohan2022rt1,mandlekar2023mimicgen,jiang2025dexmimicgen,wu2024gello,khazatsky2024droid,openx2023}.
For contact-rich manipulation, force-centered imitation work further shows that demonstrations encode contact skills, not only motion~\citep{xie2024justaddforce,yu2024mimictouch,liu2025forcemimic,zhang2025kinedex,liu2026dexteleop0forceawarebimanualdexterous}.
\name builds on this view by adapting human-guided trajectories when robot-side fingertip force disagrees with the intended contact.

\paragraph{Dexterous retargeting.}
Dexterous hand retargeting seeks to transfer human hand motion to robot hands despite differences in morphology, actuation, sensing, and compliance.
Teleoperation, human-video, mixed-reality, and portable motion-capture systems improve demonstration collection, while reinforcement learning, residual refinement, and human-like hardware reduce embodiment mismatch~\citep{handa2020dexpilot,qin2022dexmv,ye2023learning,sivakumar2022robotictelekinesis,qin2023anyteleop,arunachalam2023holodex,wang2024dexcap,zhao2024dexh2r,liu2025dextrack,xin2025retargetingobjectives,ye2025power,liu2026dexteleop0forceawarebimanualdexterous}.
These methods primarily obtain, map, or refine human motion; \name instead studies how a robot hand should adapt a human-guided motion and force reference under real-time contact feedback.

\paragraph{Force and tactile feedback for contact-rich manipulation.}
Hybrid position/force and impedance control provide reactive contact regulation, but require task-specific gains and hand-designed force-to-motion mappings~\citep{raibert1981hybrid,hogan1985impedance,liang2023hfvc}.
Learning-based tactile manipulation and recent visual-tactile policies show that contact feedback improves grasp adjustment, in-hand dexterity, and contact-rich policies when vision and proprioception are insufficient~\citep{kappassov2015tactilereview,yuan2017gelsight,calandra2018more,lambeta2020digit,bhirangi2021reskin,yin2023touchdexterity,qi2023rotateit,yuan2023robotsynesthesia,higuera2024sparsh,xue2025reactivediffusion}.
\name keeps this fast local feedback role, but learns a retargeting controller that can be composed with multiple human-guided trajectory sources.


\section{Method}
\label{sec:method}

\begin{figure}[t]
    \centering
    \includegraphics[width=\linewidth]{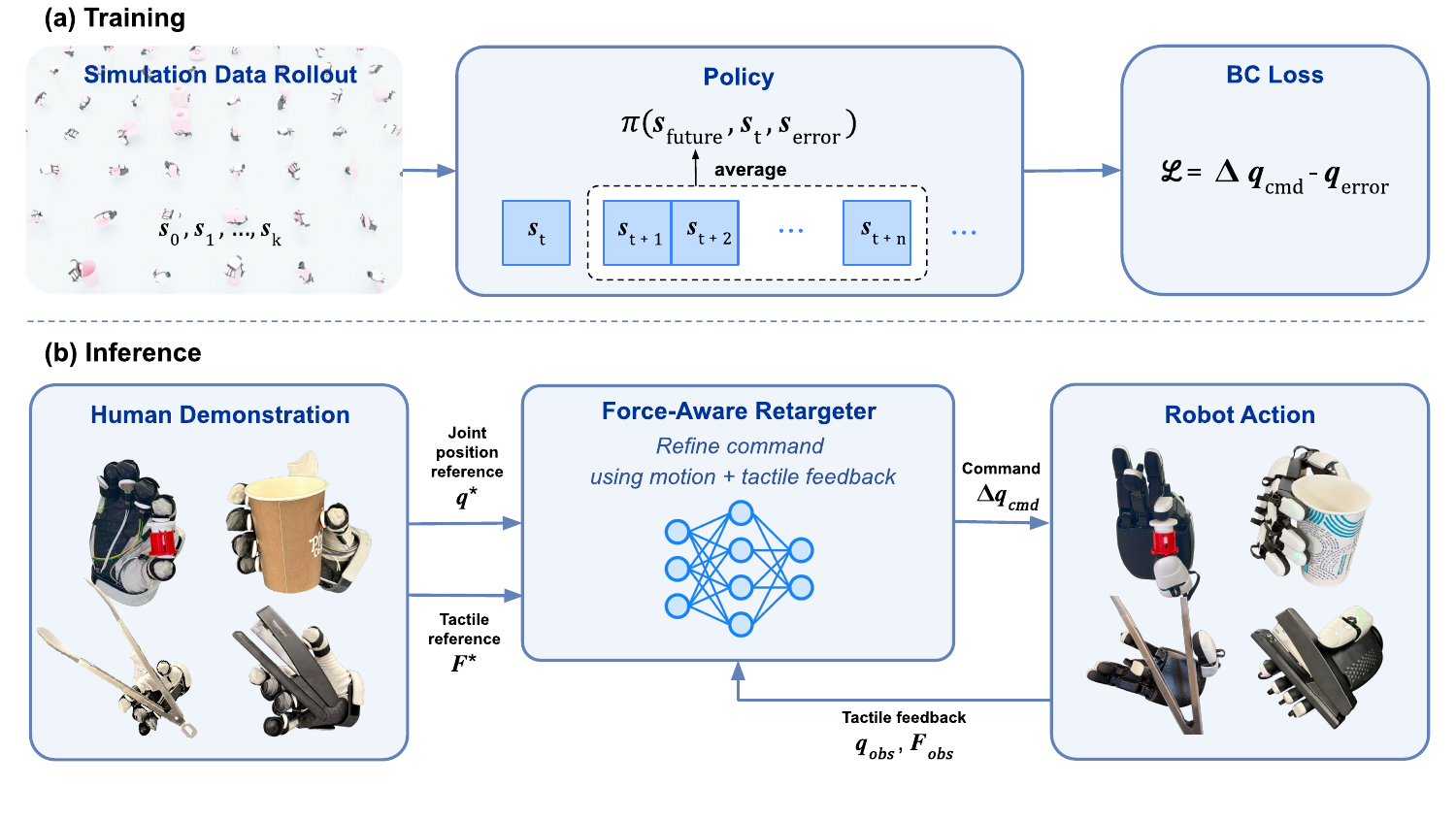}
    \caption{\textbf{ReForce training and inference pipeline.} During training, randomized and augmented simulation interactions provide joint and force trajectories from which future-window motion and contact targets are constructed. During deployment, \name adapts references using online fingertip force feedback.}
    \label{fig:pipeline}
\end{figure}

\subsection{Problem Formulation}

The \name policy is designed as a closed-loop force-aware retargeting
controller that adapts an upstream motion-and-contact reference using the
current hand state and fingertip force feedback. 
At control step $t$, let
$q_t^{\mathrm{obs}}\in\mathbb{R}^{D_q}$ denote the observed hand joint
configuration and $F_t\in\mathbb{R}^{D_F}$ the measured fingertip
normal forces, where $D_q$ is degrees of freedom and $D_F=5$
corresponds to the thumb, index, middle, ring, and pinky. The measured
state, upstream target, and target--measurement error are
\begin{equation}
    s_t
    =
    \left(q_t^{\mathrm{obs}},F_t\right),
    \qquad
    s_t^\star
    =
    \left(q_t^\star,F_t^\star\right),
    \qquad
    e_t
    =
    \left(
    q_t^\star-q_t^{\mathrm{obs}},
    F_t^\star-F_t
    \right).
\end{equation}
Here, $s_t^\star$ is the reference motion-and-contact state supplied by an
upstream reference source, where $q_t^\star$ denotes the target joint
configuration and $F_t^\star$ denotes the target contact force.
Before being passed to the network, all input quantities are standardized. The standardization procedure is described in
Appendix~\ref{sec:appendix_standard}.
The \name policy, parameterized by $\theta$, predicts an bounded joint-command update.
\begin{equation}
    {\Delta q}_t^{\mathrm{cmd}}
    =
    \pi_\theta
    \left(
    s_t,
    s_t^\star,
    e_t
    \right).
\end{equation}
Let $q_t^{\mathrm{cmd}}$ denote the accumulated joint command at control step $t$. The bounded update is accumulated and clipped to the hardware joint limits:
\begin{equation}
    q_{t+1}^{\mathrm{cmd}}
    =
    \operatorname{clip}
    \left(
    q_t^{\mathrm{cmd}}
    +
    \Delta q_t^{\mathrm{cmd}},
    q_{\min},
    q_{\max}
    \right).
\end{equation}
Here, $q_{\min},q_{\max}\in\mathbb{R}^{D_q}$ are the elementwise lower
and upper joint limits, respectively.

\subsection{Force-Aware Policy Learning and Execution}
\label{sec:reforce_policy}

\paragraph{Post-hoc motion-and-contact targets.}
The policy is trained from simulated hand--object interaction trajectories
$\{(q_t,F_t)\}_{t=1}^{T}$ collected under randomized
hand configurations, motion trajectories, and object contact
properties. The post-hoc motion and force targets are
\begin{equation}
    q_t^\star
    =
    \frac{1}{K_t}
    \sum_{k=1}^{K_t}
    q_{t+k},
    \qquad
    F_t^\star
    =
    \frac{1}{K_t}
    \sum_{k=1}^{K_t}
    F_{t+k},
\end{equation}
where $K_t$ is the number of available future steps in the target windows. These targets summarize a nearby motion-and-contact state reached later in
the rollout.
The supervised action is instead the demonstrated one-step joint
update,
$
    \Delta q_t^{\mathrm{demo}}
    =
    q_{t+1}-q_t,
$
so the post-hoc target provides future motion-and-contact context rather
than serving directly as the action label.

\paragraph{Behavior-cloning objective.}
Let $i\in\{1,\ldots,5\}$ index the finger groups and let
$D_i$ be the number of joints associated with finger $i$. Let
$\mathcal{D}$ denote the training-sample distribution. \name is trained
using mean-squared error, first averaged over the joints within each finger
and then averaged equally across fingers:
\begin{equation}
    \mathcal{L}_{\mathrm{ReForce}}
    =
    \mathbb{E}_{t\sim\mathcal{D}}
    \left[
        \frac{1}{5}
        \sum_{i=1}^{5}
        \frac{1}{D_i}
        \left\|
        \Delta q_t^{\mathrm{demo},(i)}
        -
        \Delta q_t^{\mathrm{cmd},(i)}
        \right\|_2^2
        \right].
\end{equation}
The training data include variations in contact
timing, force response, missing contact, and residual force, exposing
the controller to contact deviations that may also occur during
real-world deployment.

\subsection{Learning from Human Demonstration}
\label{sec:learn_from_human}

\paragraph{Human demonstration retargeting.}
Human hand motion is retargeted into robot-hand joint configurations,
while tactile measurements are calibrated
and mapped to the corresponding robot fingertips. This produces the
paired reference trajectory
$
    \tau^{\mathrm{ref}}
    =
    \left\{
    \left(
    q_t^{\mathrm{ref}},
    F_t^{\mathrm{ref}}
    \right)
    \right\}_{t=1}^{T}
$,
where $T$ is the number of demonstration frames,
$q_t^{\mathrm{ref}}$ is the retargeted robot configuration, and
$F_t^{\mathrm{ref}}$ is the corresponding per-finger force reference.

\paragraph{Action-chunking policy.}
We train an ACT policy~\citep{zhao2023aloha}, parameterized by
$\eta$, to predict a short chunk of future robot-space motion-and-force
references. Its configuration-history input is
$
    \mathbf{Q}_t
    =
    \left(
    q_{t-H+1},\ldots,q_t
    \right)
$,
where $H$ is the history length. During training, this history is drawn
from the retargeted demonstration trajectory; during deployment, it is
updated from the configured runtime history source.

The policy contains separate motion and force prediction branches. The
motion branch conditions on the configuration history and task phase $\phi_t$,
whereas the force branch conditions on the task phase alone:
\begin{equation}
    \left\{
    \hat q_{t+k\mid t}^{\mathrm{ref}}
    \right\}_{k=0}^{C-1}
    =
    \pi_\eta^q
    \left(
    \mathbf{Q}_t,\phi_t
    \right),
    \qquad
    \left\{
    \hat F_{t+k\mid t}^{\mathrm{ref}}
    \right\}_{k=0}^{C-1}
    =
    \pi_\eta^F
    \left(
    \phi_t
    \right).
\end{equation}
Here, $C$ is the chunk length, $k$ is the offset within the chunk, and
$\hat q_{t+k\mid t}^{\mathrm{ref}}$ denotes the reference for time
$t+k$ predicted from the policy input at time $t$. The model is
deterministic and does not take the measured fingertip force $F_t$ as
input. Consequently, the predicted force chunk represents phase-dependent
contact intent learned from the demonstrations, while \name uses online
force feedback to adapt the robot command.

\paragraph{Training objective.}

Using uniform temporal weighting, the motion and force prediction
losses for a complete chunk are
\begin{equation}
    \mathcal{L}_{q}^{\mathrm{MSE}}
    =
    \frac{1}{C D_q}
    \sum_{k=0}^{C-1}
    \left\|
    \hat q_{t+k\mid t}^{\mathrm{ref}}
    -
    q_{t+k}^{\mathrm{ref}}
    \right\|_2^2,
    \qquad
    \mathcal{L}_{F}^{\mathrm{MSE}}
    =
    \frac{1}{C D_F}
    \sum_{k=0}^{C-1}
    \left\|
    \hat F_{t+k\mid t}^{\mathrm{ref}}
    -
    F_{t+k}^{\mathrm{ref}}
    \right\|_2^2.
\end{equation}
The total training objective is
\begin{equation}
    \mathcal{L}_{\mathrm{ACT}}
    =
    \mathcal{L}_{q}^{\mathrm{MSE}}
    +
    \lambda_F
    \mathcal{L}_{F}^{\mathrm{MSE}},
\end{equation}
where $\lambda_F\geq 0$ controls the relative weight of the force
prediction loss. Invalid terminal-padding positions are masked out
before averaging.

\paragraph{Composition with \name.}

At deployment, the action-chunking policy is queried periodically to
produce motion-and-force reference chunks. We apply standard temporal
ensembling over overlapping chunks: at each control step, predictions
from all active chunks that correspond to the current time are combined,
with slightly larger weights assigned to more recent policy queries.
The resulting motion reference $q_t^\star$ and force reference
$F_t^\star$ are supplied directly to \name. The measured fingertip force
is provided separately to \name as online feedback, allowing it to adapt
the robot command while tracking the predicted references.


\section{Experimental Evaluation}
\label{sec:experiment}

We evaluate \name along three questions:
(1) whether it improves real-world force tracking over direct replay and
admittance control under the same reference trajectory;
(2) whether it improves contact-aware execution when composed with a learned
reference policy; and
(3) how its training-data composition and input representation affect
performance in simulation.

\subsection{Real-World Evaluation}
\label{sec:real_world_evaluation}

\begin{figure}[t]
    \centering
    \includegraphics[width=\linewidth]{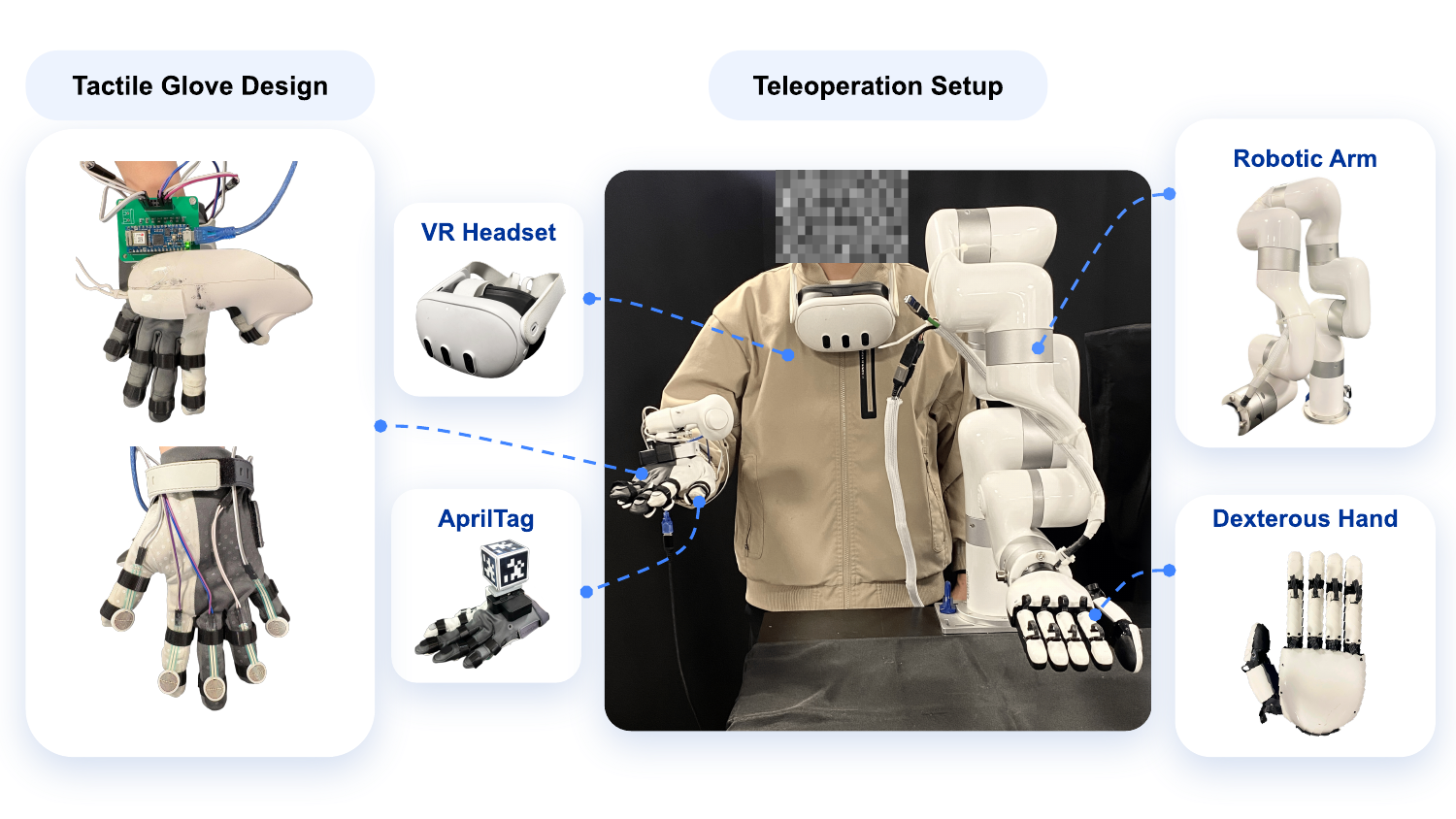}
    \caption{\textbf{Human demonstration and robot execution setup.}
        Teleoperation uses a Quest controller for wrist pose, a Manus glove for
        hand pose, and calibrated fingertip FSR sensors for human contact force.
        For ACT demonstration collection, a RealSense L515 camera tracks an
        AprilTag attached near the wrist, while the glove and FSR setup remains
        unchanged.
        A UFACTORY xArm equipped with an XHand executes the robot-side motion and
        provides fingertip tactile feedback.}
    \label{fig:hardware}
\end{figure}

\paragraph{Hardware and human demonstrations.}
Figure~\ref{fig:hardware} summarizes the sensing and execution setup.
The robot platform consists of a UFACTORY xArm equipped with a 12-DoF XHand,
whose fingertip tactile sensors provide the measured forces $F_t$.
For teleoperation, a Quest controller provides the wrist pose, a Manus glove
captures the human hand pose, and five fingertip FSR sensors provide the force
reference $F_t^\star$.
The human-side FSR measurements and robot-side tactile measurements are
calibrated in Newtons so that $F_t^\star$ and $F_t$ share a common physical
scale.

For demonstrations used to train the action-chunking reference policy, an
AprilTag and RealSense L515 RGB-D camera track the wrist alongside the glove
and fingertip-force measurements.
The policy used here is trained on the retargeted XHand joint trajectory and
the calibrated per-finger forces; the wrist trajectory is recorded
separately.

\paragraph{Tasks and evaluation criteria.}
The real-world experiments include paper-cup grasping and tongs manipulation.
The replay comparison uses side and top paper-cup grasps, while the
learned-reference evaluation includes both tasks.
These tasks require sufficient multi-finger contact while avoiding forces
that deform the manipulated object or exceed a safe operating range.

For all experiments, we measure force-tracking performance using the mean
absolute force error over time and fingertips.
For trajectory $e$ with $T_e$ logged time steps, the error is
\begin{equation}
    \mathcal{E}_{F}^{(e)}
    =
    \frac{1}{D_F T_e}
    \sum_{t=1}^{T_e}
    \sum_{i=1}^{D_F}
    \left|
        F_{e,t,i}
        -
        F_{e,t,i}^{\mathrm{ref}}
    \right|,
    \label{eq:force_tracking_error}
\end{equation}
where $F_{e,t,i}$ and $F_{e,t,i}^{\mathrm{ref}}$ are the measured and
reference forces, respectively, for fingertip $i$ at time step $t$ in
trajectory $e$.
Errors are first computed independently for each trajectory and are then
averaged equally across trials or episodes. This normalization makes the
metric independent of trajectory duration and reports force-tracking error
in Newtons.

For learned-reference experiments, we additionally report force-safe success,
the number of over-force trials, severe missing-contact trials, and the mean
number of active fingers.
For paper-cup grasping, the force threshold is $1.0$~N, while for tongs
manipulation it is $3.0$~N.
These thresholds are empirically chosen based on repeated real-world trials
to reflect task-specific force levels that avoid visible deformation or
unsafe contact.
Force-safe success requires task completion, forces below the task-specific
threshold, and no severe missing-contact failure.
A severe missing-contact failure is recorded when three or more fingers never
establish contact during the trial.
One paper-cup \name trial with missing tactile data is excluded from
force-based statistics.

\paragraph{Baselines.}
The \emph{replay} baseline directly executes the nominal joint reference
without force-dependent correction.
The \emph{admittance} baseline uses the same force reference and tactile
observations as \name, but maps force error to joint correction through a
hand-designed task-space controller. For finger $i$, it computes
\begin{equation}
    e^F_{i,t}
    =
    F^\star_{i,t}-F_{i,t},
    \qquad
    u_{i,t}
    =
    k_p e^F_{i,t}
    +
    k_d \dot e^F_{i,t}.
\end{equation}
Here, $e^F_{i,t}$ and $u_{i,t}$ are the scalar force error and force command,
$k_p$ and $k_d$ are the corresponding gains, and $\dot e^F_{i,t}$ is the
force-error derivative.
A virtual task-space admittance model converts the force command into a
fingertip displacement:
\begin{equation}
    M_i\ddot{x}_{i,t}
    +
    B_i\dot{x}_{i,t}
    +
    K_i
    \left(
    x_{i,t}-x_{i,t}^{\mathrm{ref}}
    \right)
    =
    u_{i,t}\hat{n}_{i,t}.
\end{equation}
Here, $x_{i,t}$ and $x_{i,t}^{\mathrm{ref}}$ are the virtual and reference
fingertip positions, $\hat n_{i,t}$ is the unit contact normal, and
$M_i$, $B_i$, and $K_i$ are the virtual mass, damping, and stiffness.
With $\Delta x_{i,t}=x_{i,t}-x_{i,t}^{\mathrm{ref}}$, the joint correction
is
\begin{equation}
    \Delta q^{\mathrm{adm}}_{i,t}
    =
    J_{i,t}^{\top}
    \left(
    J_{i,t}J_{i,t}^{\top}
    +
    \lambda_J^2 I_3
    \right)^{-1}
    \Delta x_{i,t}.
\end{equation}
Here, $J_{i,t}\in\mathbb{R}^{3\times D_i}$ is the fingertip Jacobian,
$\lambda_J>0$ is the damping coefficient, $I_3$ is the identity matrix, and
$\Delta q^{\mathrm{adm}}_{i,t}\in\mathbb{R}^{D_i}$ is the finger-joint
correction.
\name replaces this hand-designed force-to-motion mapping with the learned
force-aware correction policy described in
Section~\ref{sec:reforce_policy}.

\subsubsection{Force Tracking with Replayed References}
\label{sec:replay_force_tracking}

\noindent
\begin{minipage}[t]{0.40\linewidth}
    \vspace{0pt}
    Table~\ref{tab:paper_cup_replay} compares direct replay,
    admittance control, and \name, with each method driven by the
    same recorded source trajectory for a given paper-cup grasp.
    \name achieves the lowest force-tracking error for both side and
    top grasps.
\end{minipage}
\hfill
\begin{minipage}[t]{0.57\linewidth}
    \vspace{0pt}
    \centering
    \setlength{\abovecaptionskip}{0pt}
    \setlength{\belowcaptionskip}{3pt}

    \captionof{table}{\textbf{Real-world paper cup grasping replay.}
        Force-tracking error in Newtons, reported as mean $\pm$
        standard deviation over five trials. Lower is better.}
    \label{tab:paper_cup_replay}

    \small
    \setlength{\tabcolsep}{3pt}
    \resizebox{\linewidth}{!}{%
        \begin{tabular}{lccc}
            \toprule
            \textbf{Grasp}
            &
            \textbf{Replay}
            &
            \textbf{Admittance}
            &
            \name
            \\
            \midrule
            Side
            &
            $0.309 \pm 0.006$
            &
            $0.280 \pm 0.013$
            &
            $\mathbf{0.247 \pm 0.036}$
            \\
            Top
            &
            $0.736 \pm 0.000$
            &
            $0.619 \pm 0.048$
            &
            $\mathbf{0.474 \pm 0.049}$
            \\
            \bottomrule
        \end{tabular}%
    }
\end{minipage}

\vspace{0.5em}

\begin{figure}[t]
    \centering
    \includegraphics[width=0.9\linewidth]{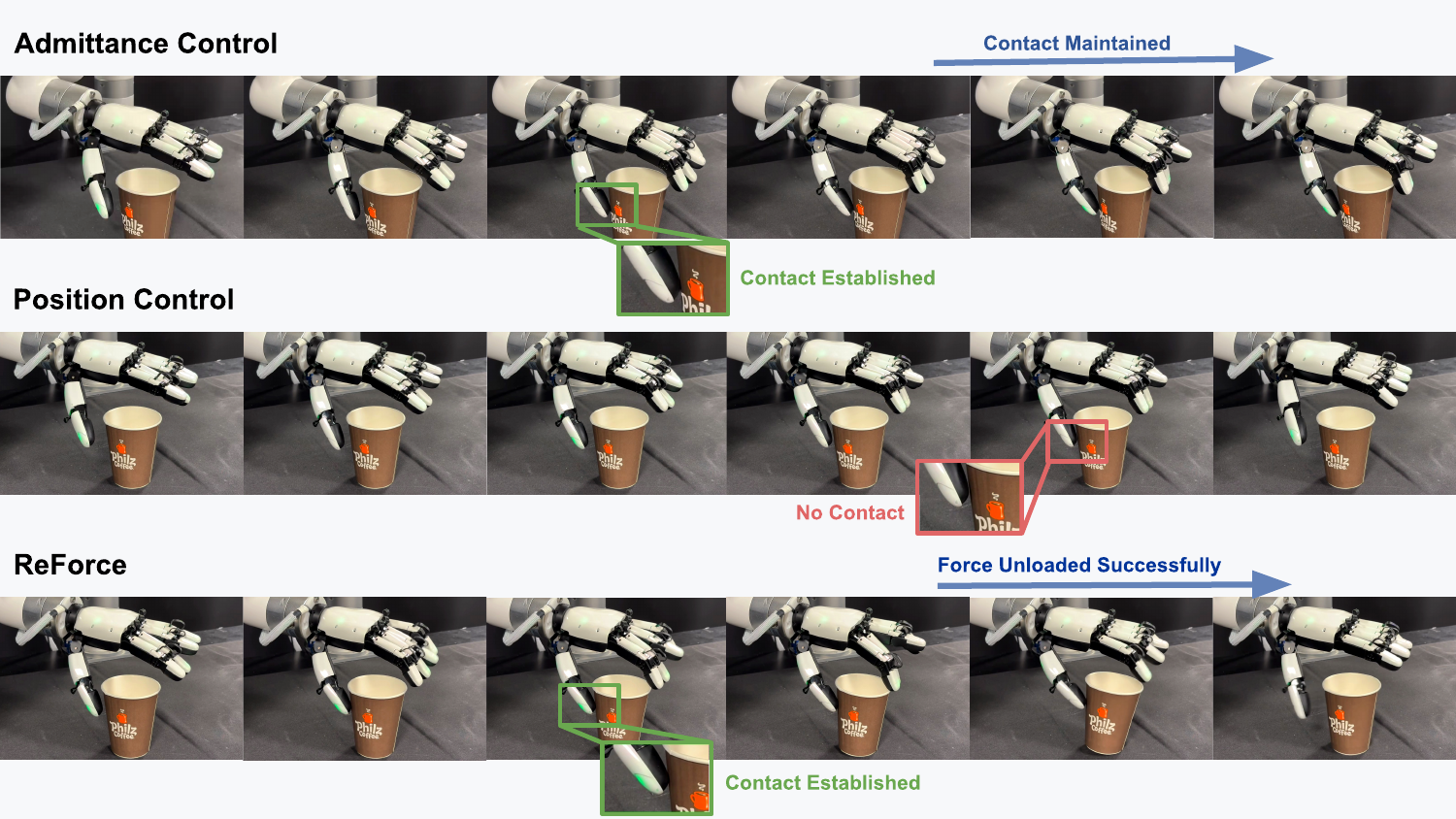}
    \caption{\textbf{Qualitative comparison of execution behavior.}
        Direct replay does not react to contact error, while admittance control
        uses a fixed force-to-motion mapping.
        \name learns contact-dependent command corrections from interaction
        data.}
    \label{fig:baseline_comparison}
\end{figure}

Figure~\ref{fig:baseline_comparison} summarizes the qualitative differences
among the three execution methods.
Detailed force trajectories for both grasp configurations are provided in
Appendix~\ref{sec:appendix_force_curves},
Figure~\ref{fig:force_curves}.
Direct replay may fail to reproduce the demonstrated contact despite
following the recorded motion, while admittance control can retain residual
force after the reference decreases.
\name better balances contact establishment and force release by adapting
the command from the measured interaction state.

\subsubsection{Execution with Learned Motion-and-Force References}
\label{sec:learned_reference_execution}

We next evaluate \name when the reference is generated by the
action-chunking policy described in Section~\ref{sec:learn_from_human}.
We compare the reference policy alone, with admittance control, and with
\name.
Because low measured force can indicate either safe execution or missing
contact, we report both force-safety and contact-engagement metrics.

\begin{table*}[t]
    \centering
    \caption{\textbf{Real-world execution with learned references.}
        Force-safe success requires task completion, forces below the
        task-specific threshold, and no severe missing-contact failure.
        Tracking error is the mean absolute force error over time and
        fingertips, reported in Newtons.}
    \label{tab:il_multimetric}
    \small
    \setlength{\tabcolsep}{4pt}
    \resizebox{\linewidth}{!}{%
        \begin{tabular}{llccccc}
            \toprule
            \textbf{Task}
             &
            \textbf{Method}
             &
            \textbf{Force-safe success} $\uparrow$
             &
            \textbf{Tracking error} $\downarrow$
             &
            \textbf{Over-force} $\downarrow$
             &
            \textbf{Severe missing contact} $\downarrow$
             &
            \textbf{Active fingers} $\uparrow$
            \\
            \midrule
            Paper cup
             &
            Reference policy
             &
            $30.0\%$
             &
            $0.812$
             &
            $0/10$
             &
            $7/10$
             &
            $0.17$
            \\
            Paper cup
             &
            Reference policy + Admittance
             &
            $60.0\%$
             &
            $0.683$
             &
            $4/10$
             &
            $1/10$
             &
            $1.26$
            \\
            Paper cup
             &
            Reference policy + \name
             &
            $\mathbf{70\%}$
             &
            $\mathbf{0.124}$
             &
            $2/10$
             &
            $\mathbf{0/10}$
             &
            $\mathbf{2.61}$
            \\
            \midrule
            Tongs
             &
            Reference policy
             &
            $50.0\%$
             &
            $0.454$
             &
            $0/10$
             &
            $5/10$
             &
            $0.94$
            \\
            Tongs
             &
            Reference policy + Admittance
             &
            $\mathbf{90.0\%}$
             &
            $\mathbf{0.419}$
             &
            $0/10$
             &
            $1/10$
             &
            $1.34$
            \\
            Tongs
             &
            Reference policy + \name
             &
            $54.5\%$
             &
            $0.441$
             &
            $4/11$
             &
            $1/11$
             &
            $\mathbf{1.62}$
            \\
            \bottomrule
        \end{tabular}
    }
\end{table*}

For paper-cup grasping, \name attains the highest force-safe success,
the lowest force-tracking error, no severe missing-contact failures,
and the largest mean number of active fingers.
For tongs manipulation, admittance control achieves the lowest
force-tracking error and the highest force-safe success, while \name
achieves the largest mean number of active fingers and substantially
reduces severe missing-contact failures relative to the reference policy.
Although \name slightly improves tracking error over the reference policy
on tongs, its four over-force trials reduce force-safe success.
These results indicate that \name consistently improves contact engagement,
while its force-tracking advantage over admittance control is task-dependent.

\subsection{Simulation Ablation of Training Data and Policy Inputs}
\label{sec:simulation_policy_ablation}

We conduct a simulation ablation to study how training-data construction and policy input representation affect force tracking. Each configuration is trained with three random seeds and evaluated on the same held-out set of $N=14{,}605$ simulated episodes. We use the force-tracking metric defined in Eq.~\ref{eq:force_tracking_error}, computing the error independently for each episode and then averaging equally across all episodes. For this evaluation, $F_{e,t,i}^{\mathrm{ref}}$ is the instantaneous force $F_{e,t,i}^{\mathrm{demo}}$ from the corresponding simulated demonstration trajectory, which is distinct from the future-window target $F_t^\star$ supplied to the policy.

\paragraph{Training-data construction.}
The object assets and target grasps used to seed these demonstrations come
from Dex1B~\citep{ye2025dex1b}.
The \emph{Base} distribution contains nominal simulated demonstrations with
the standard observation perturbations used during training: Gaussian joint
noise, force noise, and per-finger force dropout.
It contains no additional structured recovery examples.
We compare it with two targeted data constructions.

\emph{Reference-stall augmentation} replaces $20\%$ of the Base samples with
examples that partially stall the pose reference by holding selected
non-exempt joint-reference components fixed for force-active fingers
($F_{t,i}^{\star}\geq0.1$~N), while retaining the demonstrated one-step
action.
This exposes the policy to stale motion references without changing the
action supervision.

\emph{Pose-drift augmentation} constructs synthetic recovery samples from
low-force states.
For each finger $i$ satisfying $F_{t,i}^{\star}<0.1~\mathrm{N}$, 
the observed joints associated with that finger are perturbed toward
joint-limit configurations.
Let $\mathcal{J}_i$ denote the joint-index set of finger $i$ and let
$\widetilde q_t^{\mathrm{obs}}$ denote the perturbed observation;
$q_{t+1}^{\mathrm{demo}}$ is the next configuration in the corresponding
simulated demonstration.
Because this transformation changes the required action, its label is
recomputed elementwise as
\begin{equation}
    \Delta q_{t,\mathcal{J}_i}^{\mathrm{recovery}}
    =
    \operatorname{clip}
    \left(
    q_{t+1,\mathcal{J}_i}^{\mathrm{demo}}
    -
    \widetilde q_{t,\mathcal{J}_i}^{\mathrm{obs}},
    -5^\circ,
    5^\circ
    \right).
\end{equation}
Only eligible fingers receive the recomputed labels; the other fingers retain
their demonstrated one-step updates.
If an episode contains no eligible low-force state, the Base sampling
procedure is used.

\begin{figure*}[t]
    \centering

    \begin{minipage}[t]{0.28\textwidth}
        \vspace{0pt}
        \centering
        \includegraphics[width=\linewidth]
        {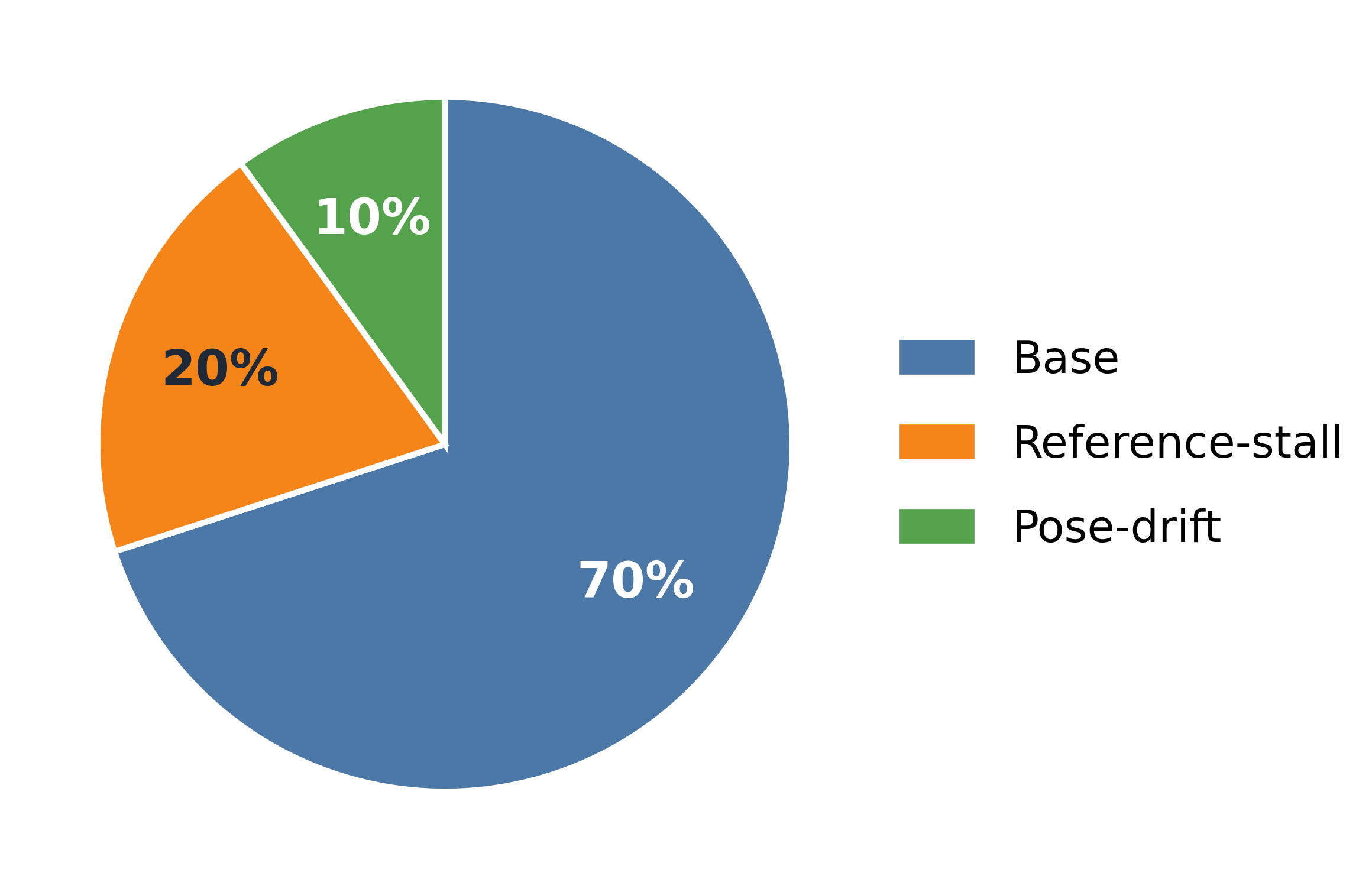}
        \captionof{figure}{
            \textbf{Composition of the training mixture.}
        }
        \label{fig:training_mixture}
    \end{minipage}
    \hfill
    \begin{minipage}[t]{0.69\textwidth}
        \vspace{0pt}
        \centering
        \captionof{table}{
            \textbf{Simulation ablation of training-data composition and
                policy inputs.}
            Force-tracking error is reported in Newtons as mean $\pm$
            sample standard deviation over three seeds.
            Lower is better.
        }
        \label{tab:simulation_policy_ablation}

        \scriptsize
        \setlength{\tabcolsep}{3pt}
        \renewcommand{\arraystretch}{1.08}

        \resizebox{\linewidth}{!}{%
            \begin{tabular}{lccc}
                \toprule
                \textbf{Training distribution}
                 &
                \shortstack{\textbf{State +}\\\textbf{force target}}
                 &
                \shortstack{\textbf{+ pose}\\\textbf{reference}}
                 &
                \shortstack{\textbf{+ error}\\\textbf{features}}
                \\
                \midrule
                Base
                 &
                $0.0601 \pm 0.0002$
                 &
                $0.0400 \pm 0.0013$
                 &
                $0.0388 \pm 0.0015$
                \\
                Base + Reference-stall aug.
                 &
                $0.0592 \pm 0.0012$
                 &
                $0.0401 \pm 0.0025$
                 &
                $0.0381 \pm 0.0005$
                \\
                Base + Pose-drift aug.
                 &
                $0.0609 \pm 0.0006$
                 &
                $0.0396 \pm 0.0010$
                 &
                $0.0385 \pm 0.0013$
                \\
                Base + Both aug.
                 &
                $0.0590 \pm 0.0012$
                 &
                $0.0396 \pm 0.0030$
                 &
                $\mathbf{0.0379 \pm 0.0006}$
                \\
                \bottomrule
            \end{tabular}%
        }
    \end{minipage}

\end{figure*}

We evaluate four training mixtures: Base alone; Base with $20\%$
reference-stall samples; Base with $10\%$ pose-drift samples; and a combined
mixture containing $70\%$ Base, $20\%$ reference-stall, and $10\%$
pose-drift samples~\ref{fig:training_mixture}.

\paragraph{Policy inputs.}
We also study how the choice of policy inputs affects force-tracking performance by comparing three progressively richer input representations:
\begin{itemize}
    \item \textbf{State and force target:}
          $\left[
                  q_t^{\mathrm{obs}},
                  F_t,
                  F_t^\star
                  \right]$;
    \item \textbf{With pose reference:}
          $\left[
                  q_t^{\mathrm{obs}},
                  F_t,
                  q_t^\star,
                  F_t^\star
                  \right]$;
    \item \textbf{With error features:}
          $\left[
                  q_t^{\mathrm{obs}},
                  F_t,
                  q_t^\star,
                  F_t^\star,
                  q_t^\star-q_t^{\mathrm{obs}},
                  F_t^\star-F_t
                  \right]$.
\end{itemize}

Adding the pose reference reduces force-tracking error across all training
distributions by approximately $32$--$35\%$ relative to the
state-and-force-target input, and explicit motion and force errors provide a
further improvement.
Both structured augmentations improve the error-feature policy individually,
while their combination performs best, attaining the lowest overall error of
$0.0379 \pm 0.0006$~N.

\section{Conclusion and Limitations}
\label{sec:conclusion}

We presented \name, a force-aware retargeting method that adapts human-guided motion and contact references using online fingertip force feedback.
In real-world experiments, \name improves force tracking and multi-finger contact engagement on contact-sensitive manipulation tasks, supporting its role as a force-aware execution layer between kinematic reference and robot control.
The current system has two main limitations. First, the ACT-style reference policy relies only on joint-state history and task phase, limiting its ability to adapt references to the observed task state; incorporating visual observations could enable more task-aware motion-and-force reference generation. Second, \name currently uses only normal fingertip force, while richer tactile signals such as shear force, slip, and contact distribution could provide a more complete representation of physical interaction.


\bibliography{reference}

@article{zhao2023aloha,
  title = {Learning Fine-Grained Bimanual Manipulation with Low-Cost Hardware},
  author = {Zhao, Tony Z. and Kumar, Vikash and Levine, Sergey and Finn, Chelsea},
  journal = {arXiv preprint arXiv:2304.13705},
  year = {2023}
}

@inproceedings{chi2023diffusionpolicy,
  title = {Diffusion Policy: Visuomotor Policy Learning via Action Diffusion},
  author = {Chi, Cheng and Feng, Siyuan and Du, Yilun and Xu, Zhenjia and Cousineau, Eric and Burchfiel, Benjamin and Song, Shuran},
  booktitle = {Robotics: Science and Systems (RSS)},
  year = {2023}
}

@article{xue2025reactivediffusion,
  title = {Reactive Diffusion Policy: Slow-Fast Visual-Tactile Policy Learning for Contact-Rich Manipulation},
  author = {Xue, Han and Ren, Jieji and Chen, Wendi and Zhang, Gu and Fang, Yuan and Gu, Guoying and Xu, Huazhe and Lu, Cewu},
  journal = {arXiv preprint arXiv:2503.02881},
  year = {2025}
}

@inproceedings{handa2020dexpilot,
  title = {{DexPilot}: Vision Based Teleoperation of Dexterous Robotic Hand-Arm System},
  author = {Handa, Ankur and Van Wyk, Karl and Yang, Wei and Liang, Jacky and Chao, Yu-Wei and Wan, Qian and Birchfield, Stan and Ratliff, Nathan D. and Fox, Dieter},
  booktitle = {IEEE International Conference on Robotics and Automation (ICRA)},
  pages = {9164--9170},
  year = {2020}
}

@article{raibert1981hybrid,
  title = {Hybrid Position/Force Control of Manipulators},
  author = {Raibert, Marc H. and Craig, John J.},
  journal = {Journal of Dynamic Systems, Measurement, and Control},
  volume = {103},
  number = {2},
  pages = {126--133},
  year = {1981}
}

@article{hogan1985impedance,
  title = {Impedance Control: An Approach to Manipulation: Part I---Theory},
  author = {Hogan, Neville},
  journal = {Journal of Dynamic Systems, Measurement, and Control},
  volume = {107},
  number = {1},
  pages = {1--7},
  year = {1985}
}

@inproceedings{yin2023touchdexterity,
  title = {Rotating without Seeing: Towards In-Hand Dexterity through Touch},
  author = {Yin, Zhao-Heng and Huang, Binghao and Qin, Yuzhe and Chen, Qifeng and Wang, Xiaolong},
  booktitle = {Robotics: Science and Systems (RSS)},
  year = {2023}
}

@article{khazatsky2024droid,
  title = {{DROID}: A Large-Scale In-The-Wild Robot Manipulation Dataset},
  author = {Khazatsky, Alexander and Pertsch, Karl and Nair, Suraj and Balakrishna, Ashwin and Dasari, Sudeep and Karamcheti, Siddharth and Nasiriany, Soroush and Srirama, Mohan Kumar and Chen, Lawrence Yunliang and Ellis, Kirsty and others},
  journal = {arXiv preprint arXiv:2403.12945},
  year = {2024}
}

@article{openx2023,
  title = {Open {X}-Embodiment: Robotic Learning Datasets and {RT-X} Models},
  author = {{Open X-Embodiment Collaboration} and others},
  journal = {arXiv preprint arXiv:2310.08864},
  year = {2023}
}

@article{liu2025forcemimic,
  title = {{ForceMimic}: Force-Centric Imitation Learning with Force-Motion Capture System for Contact-Rich Manipulation},
  author = {Liu, Wenhai and Wang, Junbo and Wang, Yiming and Wang, Weiming and Lu, Cewu},
  journal = {arXiv preprint arXiv:2410.07554},
  year = {2025}
}

@inproceedings{qin2023anyteleop,
  title = {{AnyTeleop}: A General Vision-Based Dexterous Robot Arm-Hand Teleoperation System},
  author = {Qin, Yuzhe and Yang, Wei and Huang, Binghao and Van Wyk, Karl and Su, Hao and Wang, Xiaolong and Chao, Yu-Wei and Fox, Dieter},
  booktitle = {Robotics: Science and Systems (RSS)},
  year = {2023}
}

@article{wang2024dexcap,
  title = {{DexCap}: Scalable and Portable {MoCap} Data Collection System for Dexterous Manipulation},
  author = {Wang, Chen and Shi, Haochen and Wang, Weizhuo and Zhang, Ruohan and Fei-Fei, Li and Liu, C. Karen},
  journal = {arXiv preprint arXiv:2403.07788},
  year = {2024}
}

@article{brohan2022rt1,
  title = {{RT}-1: Robotics Transformer for Real-World Control at Scale},
  author = {Brohan, Anthony and Brown, Noah and Carbajal, Justice and Chebotar, Yevgen and Dabis, Joseph and Finn, Chelsea and Gopalakrishnan, Keerthana and Hausman, Karol and Herzog, Alex and Hsu, Jasmine and others},
  journal = {arXiv preprint arXiv:2212.06817},
  year = {2022}
}

@inproceedings{arunachalam2023holodex,
  title = {{Holo-Dex}: Teaching Dexterity with Immersive Mixed Reality},
  author = {Arunachalam, Sridhar Pandian and G{\"u}zey, Irmak and Chintala, Soumith and Pinto, Lerrel},
  booktitle = {IEEE International Conference on Robotics and Automation (ICRA)},
  pages = {5962--5969},
  year = {2023}
}

@inproceedings{qin2022dexmv,
  title = {{DexMV}: Imitation Learning for Dexterous Manipulation from Human Videos},
  author = {Qin, Yuzhe and Wu, Yueh-Hua and Liu, Shaowei and Jiang, Hanwen and Yang, Ruihan and Fu, Yang and Wang, Xiaolong},
  booktitle = {European Conference on Computer Vision (ECCV)},
  year = {2022}
}

@article{zhao2024dexh2r,
  title = {{DexH2R}: Task-Oriented Dexterous Manipulation from Human to Robots},
  author = {Zhao, Shuqi and Zhu, Xinghao and Chen, Yuxin and Li, Chenran and Zhang, Xiang and Ding, Mingyu and Tomizuka, Masayoshi},
  journal = {arXiv preprint arXiv:2411.04428},
  year = {2024}
}

@article{xie2024justaddforce,
  title = {Just Add Force for Contact-Rich Robot Policies},
  author = {Xie, William and Caldararu, Stefan and Correll, Nikolaus},
  journal = {arXiv preprint arXiv:2410.13124},
  year = {2024}
}

@inproceedings{yu2024mimictouch,
  title = {{MimicTouch}: Leveraging Multi-Modal Human Tactile Demonstrations for Contact-Rich Manipulation},
  author = {Yu, Kelin and Han, Yunhai and Wang, Qixian and Saxena, Vaibhav and Xu, Danfei and Zhao, Ye},
  booktitle = {Conference on Robot Learning (CoRL)},
  year = {2024}
}

@inproceedings{higuera2024sparsh,
  title = {{Sparsh}: Self-Supervised Touch Representations for Vision-Based Tactile Sensing},
  author = {Higuera, Carolina and Sharma, Akash and Bodduluri, Chaithanya Krishna and Fan, Taosha and Lancaster, Patrick and Kalakrishnan, Mrinal and Kaess, Michael and Boots, Byron and Lambeta, Mike and Wu, Tingfan and Mukadam, Mustafa},
  booktitle = {Conference on Robot Learning (CoRL)},
  year = {2024}
}

@article{lambeta2020digit,
  title = {{DIGIT}: A Novel Design for a Low-Cost Compact High-Resolution Tactile Sensor with Application to In-Hand Manipulation},
  author = {Lambeta, Mike and Chou, Po-Wei and Tian, Stephen and Yang, Brian and Maloon, Benjamin and Most, Victoria Rose and Stroud, Dave and Santos, Raymond and Byagowi, Ahmad and Kammerer, Gregg and Jayaraman, Dinesh and Calandra, Roberto},
  journal = {IEEE Robotics and Automation Letters},
  volume = {5},
  number = {3},
  pages = {3838--3845},
  year = {2020}
}

@inproceedings{bhirangi2021reskin,
  title = {{ReSkin}: Versatile, Replaceable, Lasting Tactile Skins},
  author = {Bhirangi, Raunaq and Hellebrekers, Tess and Majidi, Carmel and Gupta, Abhinav},
  booktitle = {Conference on Robot Learning (CoRL)},
  year = {2021}
}

@article{calandra2018more,
  title = {More Than a Feeling: Learning to Grasp and Regrasp Using Vision and Touch},
  author = {Calandra, Roberto and Owens, Andrew and Jayaraman, Dinesh and Lin, Justin and Yuan, Wenzhen and Malik, Jitendra and Adelson, Edward H. and Levine, Sergey},
  journal = {IEEE Robotics and Automation Letters},
  volume = {3},
  number = {4},
  pages = {3300--3307},
  year = {2018}
}

@article{yuan2023robotsynesthesia,
  title = {Robot Synesthesia: In-Hand Manipulation with Visuotactile Sensing},
  author = {Yuan, Ying and Che, Haichuan and Qin, Yuzhe and Huang, Binghao and Yin, Zhao-Heng and Lee, Kang-Won and Wu, Yi and Lim, Soo-Chul and Wang, Xiaolong},
  journal = {arXiv preprint arXiv:2312.01853},
  year = {2023}
}

@inproceedings{shafiullah2022bet,
  title = {Behavior Transformers: Cloning $k$ Modes with One Stone},
  author = {Shafiullah, Nur Muhammad and Cui, Zichen and Altanzaya, Ariuntuya Arty and Pinto, Lerrel},
  booktitle = {Advances in Neural Information Processing Systems (NeurIPS)},
  pages = {22955--22968},
  year = {2022}
}

@inproceedings{mandlekar2023mimicgen,
  title = {{MimicGen}: A Data Generation System for Scalable Robot Learning Using Human Demonstrations},
  author = {Mandlekar, Ajay and Nasiriany, Soroush and Wen, Bowen and Akinola, Iretiayo and Narang, Yashraj and Fan, Linxi and Zhu, Yuke and Fox, Dieter},
  booktitle = {Conference on Robot Learning (CoRL)},
  pages = {1820--1864},
  year = {2023}
}

@inproceedings{jiang2025dexmimicgen,
  title = {{DexMimicGen}: Automated Data Generation for Bimanual Dexterous Manipulation via Imitation Learning},
  author = {Jiang, Zhenyu and Xie, Yuqi and Lin, Kevin and Xu, Zhenjia and Wan, Weikang and Mandlekar, Ajay and Fan, Linxi and Zhu, Yuke},
  booktitle = {IEEE International Conference on Robotics and Automation (ICRA)},
  year = {2025}
}

@inproceedings{wu2024gello,
  title = {{GELLO}: A General, Low-Cost, and Intuitive Teleoperation Framework for Robot Manipulators},
  author = {Wu, Philipp and Shentu, Yide and Yi, Zhongke and Lin, Xingyu and Abbeel, Pieter},
  booktitle = {IEEE/RSJ International Conference on Intelligent Robots and Systems (IROS)},
  pages = {12156--12163},
  year = {2024}
}

@inproceedings{sivakumar2022robotictelekinesis,
  title = {Robotic Telekinesis: Learning a Robotic Hand Imitator by Watching Humans on YouTube},
  author = {Sivakumar, Aravind and Shaw, Kenneth and Pathak, Deepak},
  booktitle = {Robotics: Science and Systems (RSS)},
  year = {2022}
}

@article{liu2025dextrack,
  title = {{DexTrack}: Towards Generalizable Neural Tracking Control for Dexterous Manipulation from Human References},
  author = {Liu, Xueyi and Adalibieke, Jianibieke and Han, Qianwei and Qin, Yuzhe and Yi, Li},
  journal = {arXiv preprint arXiv:2502.09614},
  year = {2025}
}

@article{xin2025retargetingobjectives,
  title = {Analyzing Key Objectives in Human-to-Robot Retargeting for Dexterous Manipulation},
  author = {Xin, Chendong and Yu, Mingrui and Jiang, Yongpeng and Zhang, Zhefeng and Li, Xiang},
  journal = {arXiv preprint arXiv:2506.09384},
  year = {2025}
}

@article{liu2026dexteleop0forceawarebimanualdexterous,
  title = {{DexTeleop-0}: Force-Aware Bimanual Dexterous Teleoperation with Ego-Centric Perception towards Shared Autonomy},
  author = {Haichao Liu and Yuyao Jiang and Hyunsun Park and Yuanjiang Xue and Ziwei Wang},
  journal = {arXiv preprint arXiv:2606.23431},
  year = {2026}
}

@inproceedings{zhang2025kinedex,
  title = {{KineDex}: Learning Tactile-Informed Visuomotor Policies via Kinesthetic Teaching for Dexterous Manipulation},
  author = {Zhang, Di and Yuan, Chengbo and Wen, Chuan and Zhang, Hai and Zhao, Junqiao and Gao, Yang},
  booktitle = {Conference on Robot Learning (CoRL)},
  pages = {4123--4138},
  year = {2025}
}

@inproceedings{qi2023rotateit,
  title = {General In-Hand Object Rotation with Vision and Touch},
  author = {Qi, Haozhi and Yi, Brent and Suresh, Sudharshan and Lambeta, Mike and Ma, Yi and Calandra, Roberto and Malik, Jitendra},
  booktitle = {Conference on Robot Learning (CoRL)},
  pages = {2549--2564},
  year = {2023}
}

@article{yuan2017gelsight,
  title = {{GelSight}: High-Resolution Robot Tactile Sensors for Estimating Geometry and Force},
  author = {Yuan, Wenzhen and Dong, Siyuan and Adelson, Edward H.},
  journal = {Sensors},
  volume = {17},
  number = {12},
  pages = {2762},
  year = {2017}
}

@article{kappassov2015tactilereview,
  title = {Tactile Sensing in Dexterous Robot Hands---Review},
  author = {Kappassov, Zhanat and Corrales, Juan-Antonio and Perdereau, V{\'e}ronique},
  journal = {Robotics and Autonomous Systems},
  volume = {74},
  pages = {195--220},
  year = {2015}
}

@inproceedings{liang2023hfvc,
  title = {Learning Preconditions of Hybrid Force-Velocity Controllers for Contact-Rich Manipulation},
  author = {Liang, Jacky and Cheng, Xianyi and Kroemer, Oliver},
  booktitle = {Conference on Robot Learning (CoRL)},
  pages = {679--689},
  year = {2022}
}

@inproceedings{grauman2024egoexo4d,
  title = {{Ego-Exo4D}: Understanding Skilled Human Activity from First- and Third-Person Perspectives},
  author = {Grauman, Kristen and Westbury, Andrew and Torresani, Lorenzo and Kitani, Kris and Malik, Jitendra and Afouras, Triantafyllos and Ashutosh, Kumar and Baiyya, Vijay and Bansal, Siddhant and Boote, Bikram and others},
  booktitle = {Computer Vision and Pattern Recognition (CVPR)},
  year = {2024}
}

@inproceedings{qiu2025humanpolicy,
  title = {Humanoid Policy $\sim$ Human Policy},
  author = {Qiu, Ri-Zhao and Yang, Shiqi and Cheng, Xuxin and Chawla, Chaitanya and Li, Jialong and He, Tairan and Yan, Ge and Yoon, David J. and Hoque, Ryan and Paulsen, Lars and Yang, Ge and Zhang, Jian and Yi, Sha and Shi, Guanya and Wang, Xiaolong},
  booktitle = {Conference on Robot Learning (CoRL)},
  year = {2025}
}

@inproceedings{li2025maniptrans,
  title = {{ManipTrans}: Efficient Dexterous Bimanual Manipulation Transfer via Residual Learning},
  author = {Li, Kailin and Li, Puhao and Liu, Tengyu and Li, Yuyang and Huang, Siyuan},
  booktitle = {Computer Vision and Pattern Recognition (CVPR)},
  year = {2025}
}

@inproceedings{mandi2026dexmachina,
  title = {{DexMachina}: Functional Retargeting for Bimanual Dexterous Manipulation},
  author = {Mandi, Zhao and Hou, Yifan and Fox, Dieter and Narang, Yashraj and Mandlekar, Ajay and Song, Shuran},
  booktitle = {International Conference on Machine Learning (ICML)},
  year = {2026}
}

@article{pan2025spider,
  title = {{SPIDER}: Scalable Physics-Informed Dexterous Retargeting},
  author = {Pan, Chaoyi and Wang, Changhao and Qi, Haozhi and Liu, Zixi and Bharadhwaj, Homanga and Sharma, Akash and Wu, Tingfan and Shi, Guanya and Malik, Jitendra and Hogan, Francois},
  journal = {arXiv preprint arXiv:2511.09484},
  year = {2025}
}

@article{zheng2026egoscale,
  title = {{EgoScale}: Scaling Dexterous Manipulation with Diverse Egocentric Human Data},
  author = {Zheng, Ruijie and Niu, Dantong and Xie, Yuqi and Wang, Jing and Xu, Mengda and Jiang, Yunfan and Casta{\~n}eda, Fernando and Hu, Fengyuan and Tan, You Liang and Fu, Letian and others},
  journal = {arXiv preprint arXiv:2602.16710},
  year = {2026}
}

@article{punamiya2026egoverse,
  title = {{EgoVerse}: An Egocentric Human Dataset for Robot Learning from Around the World},
  author = {Punamiya, Ryan and Kareer, Simar and Liu, Zeyi and Citron, Josh and Qiu, Ri-Zhao and Cai, Xiongyi and Gavryushin, Alexey and Chen, Jiaqi and Liconti, Davide and Zhu, Lawrence Y and others},
  journal = {arXiv preprint arXiv:2604.07607},
  year = {2026}
}

@article{kareer2025emergence,
  title = {Emergence of human to robot transfer in vision-language-action models},
  author = {Kareer, Simar and Pertsch, Karl and Darpinian, James and Hoffman, Judy and Xu, Danfei and Levine, Sergey and Finn, Chelsea and Nair, Suraj},
  journal = {arXiv preprint arXiv:2512.22414},
  year = {2025}
}

@inproceedings{grauman2022ego4d,
  title = {{Ego4D}: Around the World in 3,000 Hours of Egocentric Video},
  author = {Grauman, Kristen and Westbury, Andrew and Byrne, Eugene and Chavis, Zachary and Furnari, Antonino and Girdhar, Rohit and Hamburger, Jackson and Jiang, Hao and Liu, Miao and Liu, Xingyu and others},
  booktitle = {Computer Vision and Pattern Recognition (CVPR)},
  pages = {18995--19012},
  year = {2022}
}

@inproceedings{hou2025adaptive,
  title = {Adaptive compliance policy: Learning approximate compliance for diffusion guided control},
  author = {Hou, Yifan and Liu, Zeyi and Chi, Cheng and Cousineau, Eric and Kuppuswamy, Naveen and Feng, Siyuan and Burchfiel, Benjamin and Song, Shuran},
  booktitle = {IEEE International Conference on Robotics and Automation (ICRA)},
  pages = {4829--4836},
  year = {2025}
}

@article{adeniji2025feel,
  title = {Feel the force: Contact-driven learning from humans},
  author = {Adeniji, Ademi and Chen, Zhuoran and Liu, Vincent and Pattabiraman, Venkatesh and Bhirangi, Raunaq and Haldar, Siddhant and Abbeel, Pieter and Pinto, Lerrel},
  journal = {arXiv preprint arXiv:2506.01944},
  year = {2025}
}

@article{luo2025being,
  title = {{Being-H0}: Vision-Language-Action Pretraining from Large-Scale Human Videos},
  author = {Luo, Hao and Feng, Yicheng and Zhang, Wanpeng and Zheng, Sipeng and Wang, Ye and Yuan, Haoqi and Liu, Jiazheng and Xu, Chaoyi and Jin, Qin and Lu, Zongqing},
  journal = {arXiv preprint arXiv:2507.15597},
  year = {2025}
}

@article{ye2023learning,
  title = {Learning continuous grasping function with a dexterous hand from human demonstrations},
  author = {Ye, Jianglong and Wang, Jiashun and Huang, Binghao and Qin, Yuzhe and Wang, Xiaolong},
  journal = {IEEE Robotics and Automation Letters},
  volume = {8},
  number = {5},
  pages = {2882--2889},
  year = {2023}
}

@article{ye2025power,
  title = {From power to precision: Learning fine-grained dexterity for multi-fingered robotic hands},
  author = {Ye, Jianglong and Wei, Lai and Jiang, Guangqi and Jing, Changwei and Zou, Xueyan and Wang, Xiaolong},
  journal = {arXiv preprint arXiv:2511.13710},
  year = {2025}
}

@article{ye2025dex1b,
  title = {{Dex1B}: Learning with 1B Demonstrations for Dexterous Manipulation},
  author = {Ye, Jianglong and Wang, Keyi and Yuan, Chengjing and Yang, Ruihan and Li, Yiquan and Zhu, Jiyue and Qin, Yuzhe and Zou, Xueyan and Wang, Xiaolong},
  journal = {arXiv preprint arXiv:2506.17198},
  year = {2025}
}

@article{zhu2026chord,
  title = {Learning Dexterous Manipulation Using Contact Wrench Guidance From Human Demonstration},
  author = {Zhu, Xinghao and Liu, Zixi and Jain, Shalin and Li, Chenran and Noori, Milad and Zhao, Huihua and Welsh, John and Lin, Michael Andres and Liu, Wei and Wang, Tingwu and Da, Xingye and Luo, Zhengyi and Kulkarni, Vishal and Bhatti, Naema and Zhu, Yuke and Fan, Linxi and Wen, Bowen and Xu, Danfei and Pouya, Soha and Chang, Yan},
  journal = {arXiv preprint arXiv:2607.00033},
  year = {2026}
}
\clearpage
\onecolumn
\raggedbottom
\appendix

\setlength{\parskip}{0pt}
\setlength{\intextsep}{7pt plus 1pt minus 1pt}
\setlength{\textfloatsep}{7pt plus 1pt minus 1pt}
\setlength{\floatsep}{7pt plus 1pt minus 1pt}
\setlength{\abovecaptionskip}{4pt}
\setlength{\belowcaptionskip}{1pt}

\begin{center}
    {\Large\bfseries Appendix}
\end{center}

\section{Training Configuration}
\label{sec:appendix_training}

\subsection{\name Training Configuration}
\label{sec:appendix_reforce_training}
Table~\ref{tab:supp_hyperparameters} summarizes the main training and
architecture settings used for the \name policy in our experiments.
\begin{table}[H]
    \centering
    \caption{\textbf{\name training configuration.}
    Main hyperparameters used for the reported experiments.}
    \label{tab:supp_hyperparameters}
    \small
    \renewcommand{\arraystretch}{1.03}
    \setlength{\tabcolsep}{6pt}
    \begin{tabular}{@{}p{0.28\textwidth}p{0.64\textwidth}@{}}
        \toprule
        \textbf{Item} & \textbf{Value} \\
        \midrule

        Future-target horizon
        &
        Up to $16$ future frames; joint and force targets are averaged
        over the available future window
        \\

        Policy architecture
        &
        Per-finger MLP policy; the four non-thumb fingers use a shared
        $[128,128]$ trunk with $[64]$ finger-specific heads, while the
        thumb uses a separate $[128,128]$ trunk and $[64]$ head;
        ReLU activations
        \\

        Joint-update limit
        &
        $0.088$ rad per controlled joint, approximately $5^\circ$
        \\

        Observation perturbations
        &
        Gaussian joint noise, Gaussian force noise with standard
        deviation $0.1$~N, and per-finger force dropout with
        probability $0.30$
        \\

        Optimization
        &
        AdamW with learning rate $10^{-3}$, weight decay $10^{-5}$,
        batch size $1024$, $20$ epochs, and $500$ training steps
        per epoch
        \\

        \bottomrule
    \end{tabular}
\end{table}

\subsection{ACT Reference-Policy Configuration}
\label{sec:appendix_act_training}

A separate ACT-style reference policy is trained for each task using
approximately $30$--$40$ human demonstrations.
All reported task configurations use no visual input; the policy therefore
takes only the configuration history $\mathbf{Q}_t$ and task phase $\phi_t$
as input.

\begin{table}[H]
    \centering
    \caption{\textbf{ACT reference-policy training configuration.}
    Main parameters used for the learned motion-and-force reference policies.}
    \label{tab:supp_act_hyperparameters}
    \small
    \renewcommand{\arraystretch}{1.03}
    \setlength{\tabcolsep}{6pt}
    \begin{tabular}{@{}p{0.28\textwidth}p{0.64\textwidth}@{}}
        \toprule
        \textbf{Item} & \textbf{Value} \\
        \midrule

        Training data
        &
        Separate policy per task; approximately $30$--$40$ human
        demonstrations per task
        \\

        Temporal context
        &
        History length $H=9$ and prediction chunk length $C=32$
        \\

        Sampling frequency
        &
        Raw demonstrations at $83$~Hz; training anchors sampled at
        $10$~Hz
        \\

        Phase representation
        &
        $D_\phi=3$, with
        $\phi_t=
        [\tau_t,\sin(2\pi\tau_t),\cos(2\pi\tau_t)]$,
        where
        $\tau_t=(t-1)/(T-1)$
        \\

        Force-loss weight
        &
        $\lambda_F=1.0$
        \\

        Network architecture
        &
        Context MLP with widths $[128,128,256]$, followed by separate
        $2$-layer motion and force Transformer decoders with width $256$,
        $4$ attention heads, feed-forward width $1024$, and dropout $0.1$
        \\

        \bottomrule
    \end{tabular}
\end{table}

\subsection{Policy Input Standardization}
\label{sec:appendix_standard}
Joint configurations and forces are normalized before being used by the
reference policy.
Joint configurations are standardized directly using their training-set
statistics.
Because the force distribution is strongly concentrated near zero and has a
long positive tail, force values are first log-transformed and then
standardized.
Elementwise, the transformations are
\begin{equation}
    \bar q
    =
    \frac{q-\mu_q}{\sigma_q},
    \qquad
    \bar F
    =
    \frac{
        \log\left(1+\max(F,0)/s_F\right)-\mu_F
    }{\sigma_F},
\end{equation}
where $\mu_q$ and $\sigma_q$ are the training-set statistics of the joint
configurations, $\mu_F$ and $\sigma_F$ are the statistics of the
log-transformed forces, and $s_F=1.0$~N is the force log-scale parameter.

\section{Real-Robot Force-Tracking Curves}
\label{sec:appendix_force_curves}

\begin{figure}[H]
    \centering
    \includegraphics[width=0.90\textwidth]
    {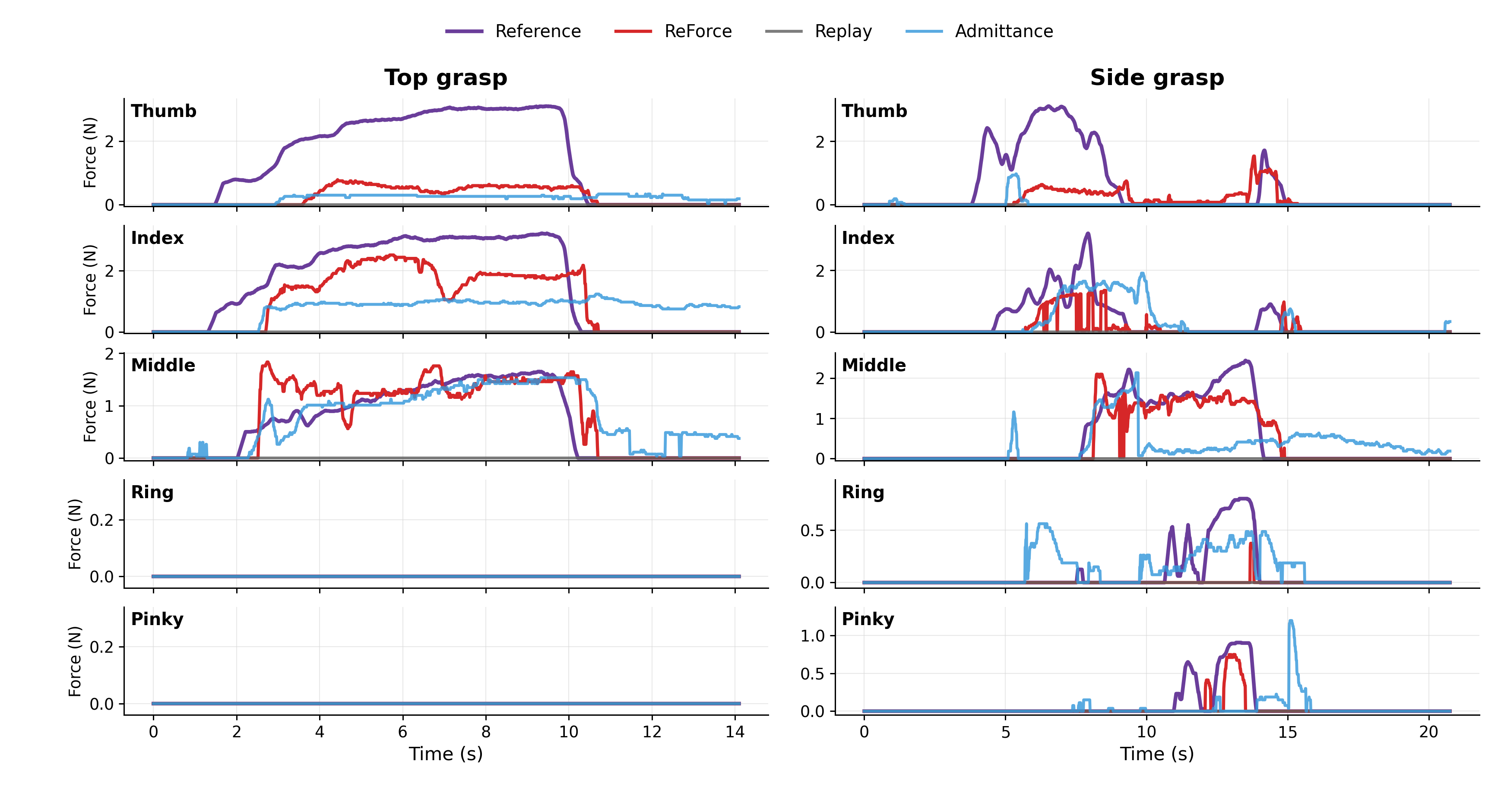}
    \caption{\textbf{Force tracking under replayed references.}
        Real-robot force trajectories for side and top paper-cup grasps using
        direct replay, admittance control, and \name.
        These curves correspond to the results in
        Table~\ref{tab:paper_cup_replay}.}
    \label{fig:force_curves}
\end{figure}

\begin{figure}[H]
    \centering

    \begin{minipage}[t]{0.31\textwidth}
        \centering
        \includegraphics[width=\linewidth]
        {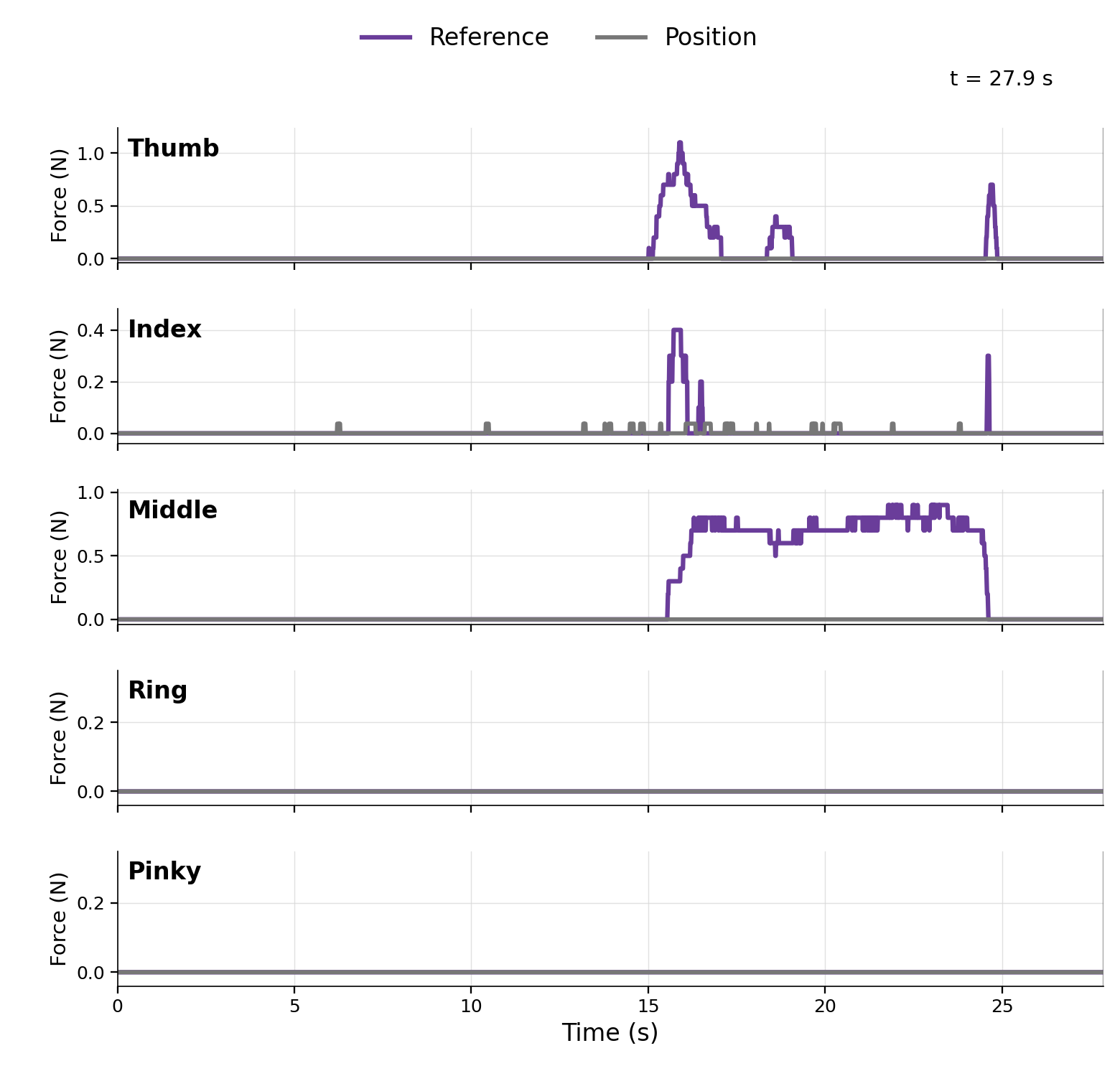}

        \vspace{-0.3em}
        \textbf{(a) Position replay}
    \end{minipage}
    \hfill
    \begin{minipage}[t]{0.31\textwidth}
        \centering
        \includegraphics[width=\linewidth]
        {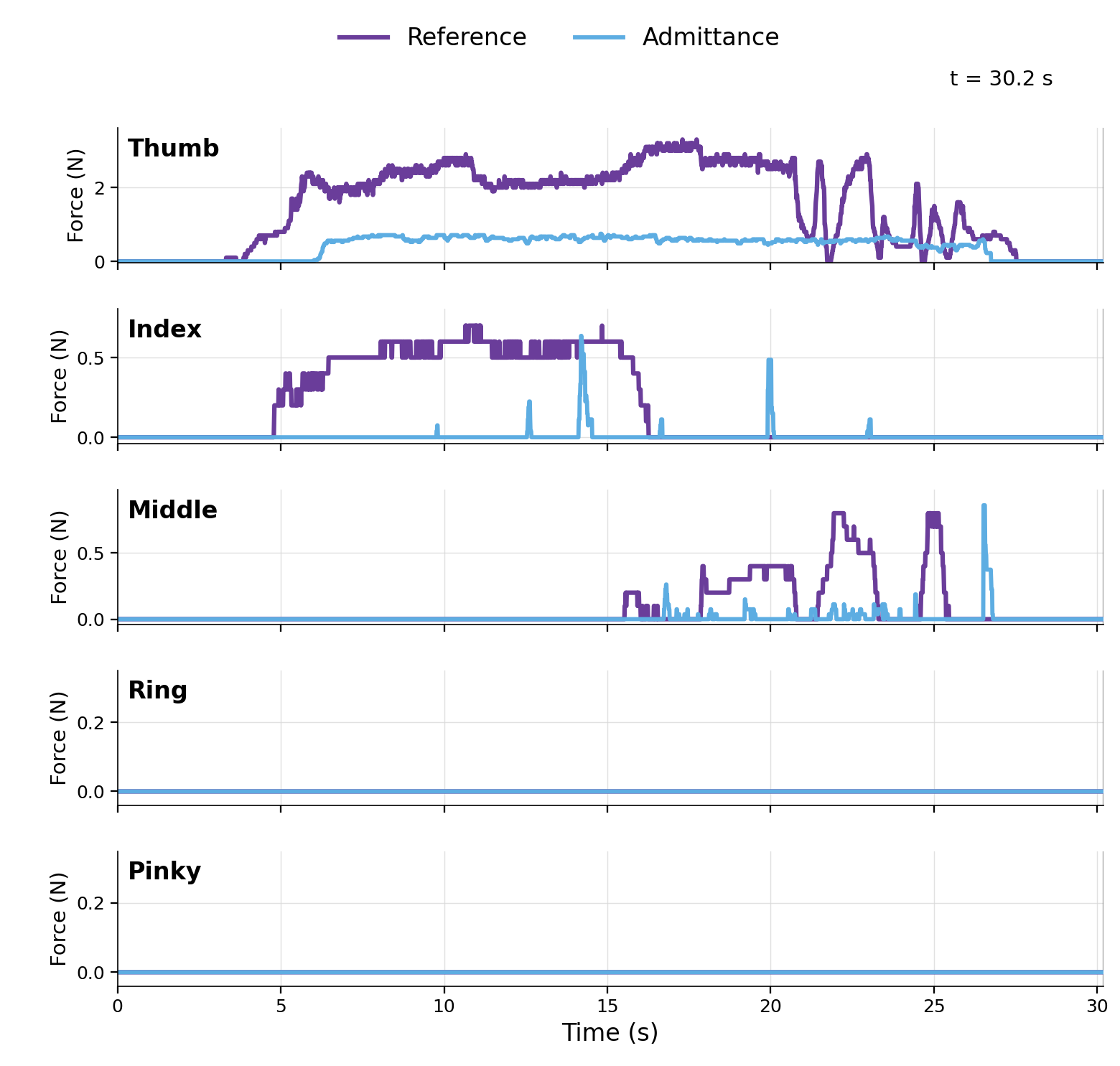}

        \vspace{-0.3em}
        \textbf{(b) Admittance control}
    \end{minipage}
    \hfill
    \begin{minipage}[t]{0.31\textwidth}
        \centering
        \includegraphics[width=\linewidth]
        {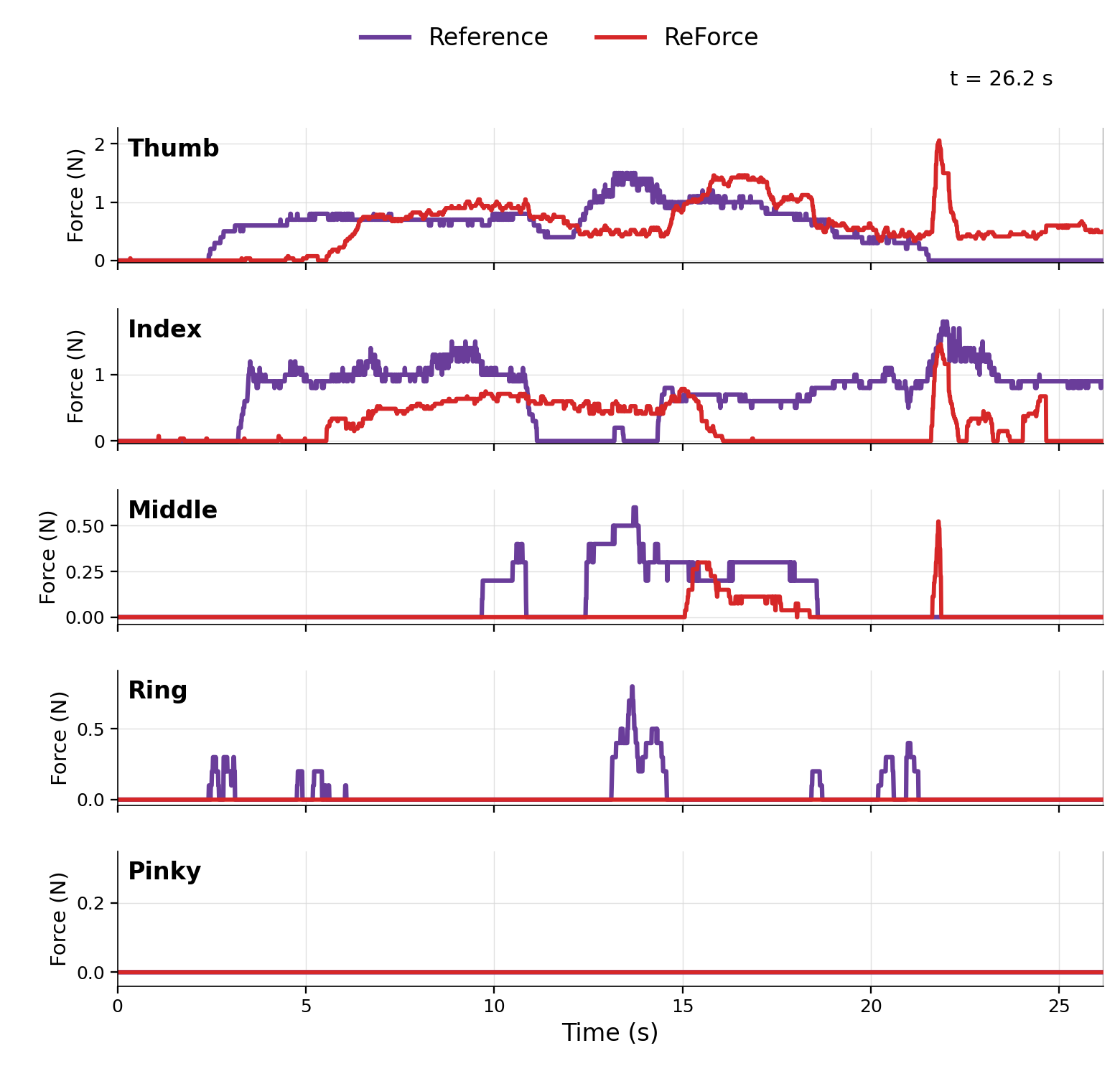}

        \vspace{-0.3em}
        \textbf{(c) ReForce}
    \end{minipage}

    \caption{\textbf{Representative paper-cup force-tracking trials.}
        Force trajectories produced by position replay, admittance control,
        and \name.}
    \label{fig:supp_papercup_force_tracking}

\end{figure}

\end{document}